\documentclass[conference]{IEEEtran}
\IEEEoverridecommandlockouts
\usepackage{cite}
\usepackage{amsmath,amssymb,amsfonts}
\usepackage{algorithmic}
\usepackage{graphicx}
\usepackage{multirow}
\usepackage{textcomp}
\usepackage[numbers]{natbib}
\usepackage{xcolor}
\usepackage{booktabs}
\usepackage[inline]{enumitem}

\usepackage{caption}
\usepackage{subcaption}
\usepackage{hyperref}
\hypersetup{
    colorlinks=true,
    linkcolor=blue,
    filecolor=magenta,      
    urlcolor=cyan,
    pdftitle={Overleaf Example},
    pdfpagemode=FullScreen,
    }
\usepackage[normalem]{ulem}

\newcommand{\spara}[1]{\smallskip\noindent{\bf #1}}

\makeatletter
\newcommand{\linebreakand}{%
  \end{@IEEEauthorhalign}
  \hfill\mbox{}\par
  \mbox{}\hfill\begin{@IEEEauthorhalign}
}
\makeatother

\def\BibTeX{{\rm B\kern-.05em{\sc i\kern-.025em b}\kern-.08em
    T\kern-.1667em\lower.7ex\hbox{E}\kern-.125emX}}
\begin{document}

\IEEEaftertitletext{\vspace{-3\baselineskip}}


\title{Practitioner-Centric Approach for Early Incident Detection Using Crowdsourced Data for Emergency Services}

\author{
\IEEEauthorblockN{Yasas Senarath}
\IEEEauthorblockA{\textit{George Mason University}, USA \\
ywijesu@gmu.edu}
\and
\IEEEauthorblockN{Ayan Mukhopadhyay}
\IEEEauthorblockA{\textit{Vanderbilt University}, USA \\
ayan.mukhopadhyay@vanderbilt.edu}
\and
\IEEEauthorblockN{Sayyed Mohsen Vazirizade}
\IEEEauthorblockA{\textit{Vanderbilt University}, USA \\
s.m.vazirizade@vanderbilt.edu}
\linebreakand
\IEEEauthorblockN{Hemant Purohit}
\IEEEauthorblockA{\textit{George Mason University}, USA \\
hpurohit@gmu.edu}
\and
\IEEEauthorblockN{Saideep Nannapaneni}
\IEEEauthorblockA{\textit{Wichita State University}, USA \\
saideep.nannapaneni@wichita.edu}
\and
\IEEEauthorblockN{Abhishek Dubey}
\IEEEauthorblockA{\textit{Vanderbilt University}, USA \\
abhishek.dubey@vanderbilt.edu}
}

\maketitle

\begin{abstract}
Emergency response is highly dependent on the time of incident reporting. Unfortunately, the traditional approach to receiving incident reports (e.g., calling 911 in the USA) has time delays. Crowdsourcing platforms such as Waze provide an opportunity for early identification of incidents. However, detecting incidents from crowdsourced data streams is difficult due to the challenges of noise and uncertainty associated with such data. Further, simply optimizing over detection accuracy can compromise spatial-temporal localization of the inference, thereby making such approaches infeasible for real-world deployment. 
This paper presents a novel problem formulation and solution approach for practitioner-centered incident detection using crowdsourced data by using emergency response management as a case-study. The proposed approach \textit{CROME} (Crowdsourced Multi-objective Event Detection) quantifies the relationship between the performance metrics of incident classification (e.g., F1 score) and the requirements of model practitioners (e.g., 1 km. radius for incident detection). First, we show how crowdsourced reports, ground-truth historical data, and other relevant determinants such as traffic and weather can be used together in a Convolutional Neural Network (CNN) architecture for early detection of emergency incidents. Then, we use a Pareto optimization-based approach to optimize the output of the CNN in tandem with practitioner-centric parameters to balance detection accuracy and spatial-temporal localization. Finally, we demonstrate the applicability of this approach using crowdsourced data from Waze and traffic accident reports from Nashville, TN, USA. Our experiments demonstrate that the proposed approach outperforms existing approaches in incident detection while simultaneously optimizing the needs for real-world deployment and usability. 
\end{abstract} 


\section{Introduction} 

Emergency response to incidents like road accidents and natural disasters is one of the most pressing problems faced by communities today. Cities resort to diverse responders like firefighters, paramedics, and police personnel to manage such incidents. Traditionally, emergency response pipelines were reactive, responding to incidents after they were reported officially. However, cities have evolved over the last few decades and have adopted \textit{smart} emergency response, which acts proactively by allocating responders in anticipation of future incidents~\cite{mukhopadhyay2020review}. 
Mukhopadhyay et al.~\cite{mukhopadhyay2020review} point out that recently, incident detection has become an essential addition to the smart emergency response pipeline. To understand the role of incident detection, consider a traffic accident on a highway. In such a situation, first responders dispatch resources only after a call is placed for aid (e.g., a 911 call in the USA). However, this mechanism can result in loss of time, which in emergencies could prove to be fatal~\cite{jaldell2017important}. 
This delay is also problematic for low severity incidents, which can lead to traffic congestion on the roads if not cleared in time. Based on our discussions with the Tennessee Department of Transportation, these kinds of incidents are often detected by the help trucks on patrol and manually by personnel in transportation management centers using cameras. This pipeline results in latency, which can be avoided if observers passing by incident sites have access to a convenient mechanism for reporting incidents (e.g., by using a single click of a button on a smartphone application). Then, crowdsourced reports can be used to \textit{detect} incidents before they are reported officially.


Using crowdsourced data for emergency response is highly non-trivial as such data is noisy and scattered in space and time. This uncertainty makes precise spatial and temporal localization challenging. While early information about incidents is critical, it is difficult to dispatch limited resources to potential incidents based on uncertain information. 
Consider accident reporting through the Waze application that allows users to press a smartphone button to report accidents. Due to the vehicle's motion and traffic, drivers and passengers often pass the area of an incident before clicking the button to report it.  
As a result, by the time the incident is reported, it typically consists of both spatial and temporal deviations from the original place and time of the incident. Therefore, an incident detection model must localize and detect the incident in the presence of noise and 
uncertainty. 

The goals of accurate detection and improved localization might appear contradictory. Indeed, it seems intuitive that the accuracy of a data-driven incident detector would degrade with an increased focus on localization. However, localization is imperative to respond to an incident; it is not possible to dispatch aid to an accident unless its precise location is known. At the same time, lowering the latency of detection is also critical. If we wait to acquire more information from various sources,
the uncertainty decreases; however, it increases latency. This paper focuses on optimizing the balance between the 
goals of accurate detection and localization.

Recently, there has been a growing interest in leveraging crowdsourced data in various public sector domains such as emergency management and transportation. 
Crowdsourced data (such as Twitter, Instagram, Waze, and Foursquare) represents another source of data that has been used for a variety of analyses such as emergency disaster response~\cite{senarath2020emergency}
and traffic management~\cite{amin2018evaluating}. Lenkei~\cite{lenkei2018crowdsourced} performed a comparative study of the accidents reported through Waze and the traffic database in Sweden (Trafikverket) and
observed that 27.5\% of the incidents in Trafikverket were detected earlier by Waze. Recently, Senarath et al.~\cite{senarath2020emergency} proposed a Bayesian information fusion approach which fused multiple crowdsourced Waze reports for early incident detection.

We note 
that prior works 
using crowdsourced data for spatial-temporal incident detection lack: 
\begin{enumerate*}
    \item a principled approach for information fusion noisy data to detect incidents that 
    can generalize across crowdsourcing platforms because existing approaches only consider simple aggregation methods of explicit features without considering impacts of uncertainty associated with report integration; 
    \item a systematic investigation of the effect of hyperparameters (e.g., quality of discretization of a continuous region of interest) on the quality and accuracy of the event detection models developed from crowdsourced data; and
    \item a flexible incident detection framework that accommodates practitioner preferences regarding incident detection accuracy and spatial-temporal localization.
\end{enumerate*}

This paper presents \textbf{CROME}: an approach that performs \textbf{Cro}wdsourced \textbf{M}ulti-Objective \textbf{E}vent Detection. We show how to combine spatial-temporal crowdsourced data with arbitrary features to facilitate data-driven modeling. Crucially, we show how to balance the trade-off between localization and accuracy. We also show that balancing this trade-off is vital and naively maximizing accuracy (or F-1 score) is not desirable in this problem domain. Through experiments performed on accident data gathered from Nashville, TN, USA, we show how our approach can detect accidents sooner than receiving 911 calls and outperforms state-of-the-art approaches. 

\section{CROME: Crowdsourced Multi-Objective Event Detection}
\label{sec}

Consider a set of spatial-temporal incidents $D$; for example, $D$ can denote a set of historical road accidents that occurred in a region of interest during a specific period. 
In practice, it is difficult to know the exact time of occurrence of such incidents. Instead, we assume that the reported time of occurrence (for example, through a 911 call) can be used as a proxy for the actual time of occurrence. Further, we assume access to a set of crowdsourced reports about the incidents, denoted by $W$. Each report $w_i \in W$ can be represented by the tuple $(t_i, k_i, l_i, r_i)$ where $t_i$, $k_i$, $l_i$, and $r_i$ denote the time at which the report was generated, longitude, latitude, and the reliability of report, respectively. The reliability of a report is typically provided by the crowdsourcing platform itself. For example, the long-term accuracy of the user who generated the report can be used as a proxy for reliability.

\textbf{Spatial Discretization}: We divide the overall spatial area of interest into a grid $G$ with square cells of length $\Delta s$. Each point in space (represented by a pair of latitude and longitude) can therefore be mapped to a unique cell in $G$. Upon mapping crowdsourced reports $W$ to $G$, each report $w_i \in W$ is 
denoted by $(t_i, x_i, y_i, r_i)$, where $x_i$ and $y_i$ represent a specific cell in $G$ 
containing the spatial coordinates represented by $k_i$ and $l_i$.

\textbf{Temporal Discretization}: Crowdsourced reports are generated in continuous time. However, incidents of interest, e.g., accidents, occur infrequently. Further, multiple crowdsourced reports are generated in response to the same incident. As a result, we aggregate crowdsourced reports temporally to aid analysis. Let $\Delta t$ denote the temporal resolution for data aggregation. We refer to $\Delta t$ as the \textit{step} time-period. The input data $W$ can be grouped in $\Delta t$ time intervals and aggregated to produce a real number using a specific summary statistic for temporal discretization. For example, the total number of reports generated in a time period in a cell or the average reliability of the reports can be used for aggregation.  

\begin{figure}[h]
\centering
\includegraphics[width=9cm]{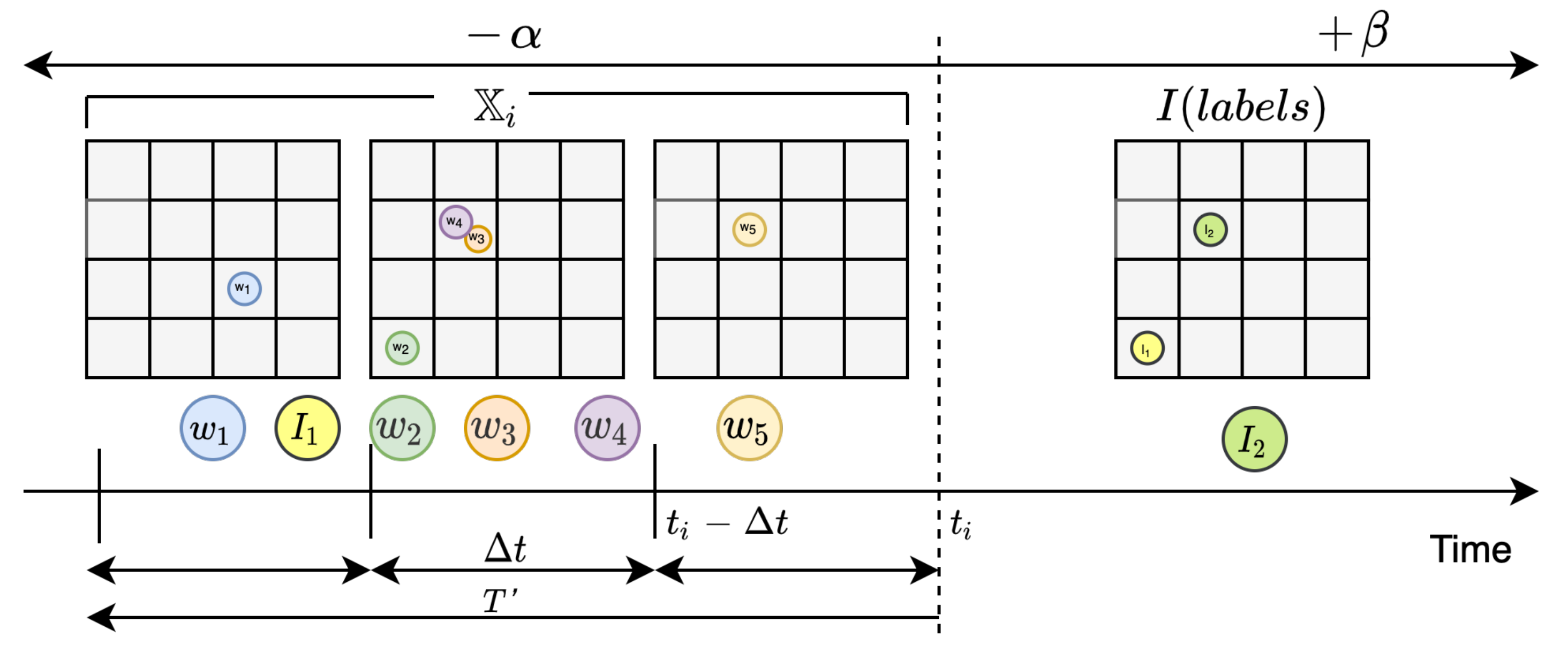}
\caption{The overall process of data discretization and aggregation. This diagram represents three time steps ($\Delta t$) aggregated to form $\mathbb{X}_i$ and the corresponding labels ($L_i$) at time $t_i$.}
\label{fig:data-aggregation}
\end{figure}

We also assume access to other potential determinants of incidents, e.g., weather and traffic congestion have been shown to be correlated with accident occurrence~\cite{mukhopadhyay2020review}. We assume that information about such covariates is available at the same granularity at which we choose to discretize space and time. 
We refer to the covariates used (including the summary statistics of the crowdsourced reports) as features. For example, features in the case of accidents can correspond to the number of crowdsourced reports, weather, congestion, and so on. 
For each time period (say a time period ending at $t_i$), using a specific 
aggregation on $W$ produces a multi-dimensional array (i.e., a tensor) $\chi_i$ of dimensions $(N_x, N_y, N_f)$, where $N_x$ and $N_y$ denote the number of cells along the horizontal and vertical directions in $G$, and $N_f$ is the number of features. We show the overall process of discretization in Figure~\ref{fig:data-aggregation}. Note that as $\Delta s$ changes, the realizations of $N_x$ and $N_y$ change as well.

In practice, once an incident occurs, crowdsourced reports are generated for some extended duration of time after the incident. Consider an accident for example; as people pass through the scene of the incident, they observe it and log the presence of the accident on a relevant crowdsourcing platform. The reporting can continue till the scene of the accident is responded to by first responders and cleared. In order to accumulate reports generated over time, we use $\mathbb{X}_{i}$ to denote $(\chi_{i-T'} \dots \chi_i)$, where $T'$ is an arbitrary period 
that practitioners can choose based on the frequency of the reports, the number of incidents, and so on. Therefore, $\mathbb{X}_{i}$ is a collection of tensors that provides a temporal snapshot of reports generated from $t_i - T'$ to $t_i$.

\textbf{Label Assignment}: Our goal is to detect incidents based on crowdsourced reports and other potential determinants. While such a problem can be framed as unsupervised anomaly detection, historical reports generated through crowdsourcing platforms and \textit{ground-truth} data about past incidents can be used to create a labeled dataset that can be used for supervised learning. Ground truth information about accidents can be obtained through historical records; for example, information about accidents can be retrieved through first responders or traffic/safety organizations. Based on such data, we assign a label to each $\mathbb{X}_{i}$ generated by accumulating crowdsourced reports. Intuitively, we say that a set of crowdsourced reports $\mathbb{X}_{i}$ \textit{match} with an actual incident (observed through ground truth historical data) if there was an actual accident in proximity of the set, for some definition of spatial-temporal proximity. 

We define proximity using two temporal hyper-parameters $\alpha$ and $\beta$ and a spatial hyper-parameter $\delta$. Each $\mathbb{X}_{i}$ (recall that $\mathbb{X}_{i}$ denotes a set of reports accumulated until time $t_i$) is assigned a positive label (marked as 1) if there was an actual incident in the time frame $t_i - \alpha$ to $t_i + \beta$ within a spatial proximity of $\delta$. It is reported a negative label (marked as 0) otherwise. On a cursory glance, it might seem non-intuitive to assign a crowdsourced report with an actual incident that might have occurred after it; 
although $\mathbb{X}_{i}$ consists of a set of reports obtained until time $t_i$ but duration $t+\beta$ might consist of incidents that occurred after it. However, our preliminary analysis revealed that in practice, it is not possible to retrieve the \textit{true} time of occurrence of incidents. For example, consider traffic accidents. Once an accident occurs, some time elapses before the victim (or an observer) calls for help and officially reports an estimated time of occurrence. Crowdsourced reports, on the other hand, can be generated as soon as the accident occurs, thereby precluding the \textit{officially reported} time of occurrence of the accident. Therefore, we use the hyper-parameter $\beta$ to accommodate such scenarios. 



\subsection{Problem Formalization}
\label{sec: Problem_Statement}

  Given an aggregation of crowdsourced reports and other relevant features, our goals are the following: \textbf{1)} early detection of the locations of spatial-temporal incidents like accidents through crowdsourced reports, and \textbf{2)} determine the best practitioner's parameters regarding localization that can be used for deployment. It is crucial to optimize parameters from the perspective of first responders and practitioners since the optimal model (based on some definition of optimality as defined in the context of data-driven learning, like maximum likelihood) might not be the best choice for deployment. For example, consider a model that detects accidents early in a city accurately most of the time, but does poorly on spatial localization. First responders would not be able to service the incident even if they are provided with an alert. Similarly, consider a pipeline that detects accidents accurately at the temporal resolution of a day; even if the pipeline shows accurate spatial localization, it does not aid emergency response because of the temporal delay it might incur. Below, we show how we formulate our objective in terms of learning performance (accuracy or some measure of it) as well as practitioner-centric parameters. 

Let the random variable $I$ denote incident occurrence that the decision-maker is unaware of ($I$ would denote the labels we generated for each $\mathbb{X}$). It is important to note that such incidents have already occurred but have not been reported when we make inferences about them; as such, the problem we are interested in is not a canonical forecasting problem. Our goal is to learn a function $f(I \mid \mathbb{X}, \theta)$, where $f$ denotes some measure of accuracy of detecting incidents $I$ conditional on features $\mathbb{X}$ and a set of parameters $\theta$. For example, $f$ could represent the F-1 score of the model or its detection accuracy. In order to balance the accuracy of detection with the temporal and spatial granularity that the model uses, we choose to maximize the accuracy of the model while simultaneously aiming for higher spatial and temporal resolutions for detection. Naturally, any approach to detection is difficult at finer spatial and temporal resolutions. Formally, we define practitioner-centric incident detection as the following multi-objective optimization problem:

\vspace{-0.21in}

\begin{equation}
\begin{aligned}
\label{eqn:optimization}
    & \max_{\theta} \gamma_1 f(I \mid \mathbb{X}, \theta) + \min_{\Delta t} \gamma_2 z_1(\Delta t) + \min_{\Delta s} \gamma_3 z_2(\Delta s)
\end{aligned}
\end{equation}
where $\gamma_1,\gamma_2,$ and $\gamma_3$ denote the relative importance of each term in the optimization problem. In the simplest case, $\gamma_1 = \gamma_2 = \gamma_3 = 1$. The functions $z_1$ and $z_2$ denote arbitrary (increasing) functions that practitioners can choose over spatial and temporal discretization. In the simplest case (the one we use later in our approach), $z_1$ and $z_2$ denote identity functions. While our problem structure is general for any type of spatial-temporal incident and crowdsourced report, we use road accidents as a case study. 
We use crowdsourced reports from Waze 
that facilitates users to upload observed accident locations.



\begin{figure*}[!h]
\centering
\begin{subfigure}[b]{0.49\textwidth}
    \centering
    \includegraphics[trim={2.5cm 4cm 0 2cm},clip,width=\textwidth]{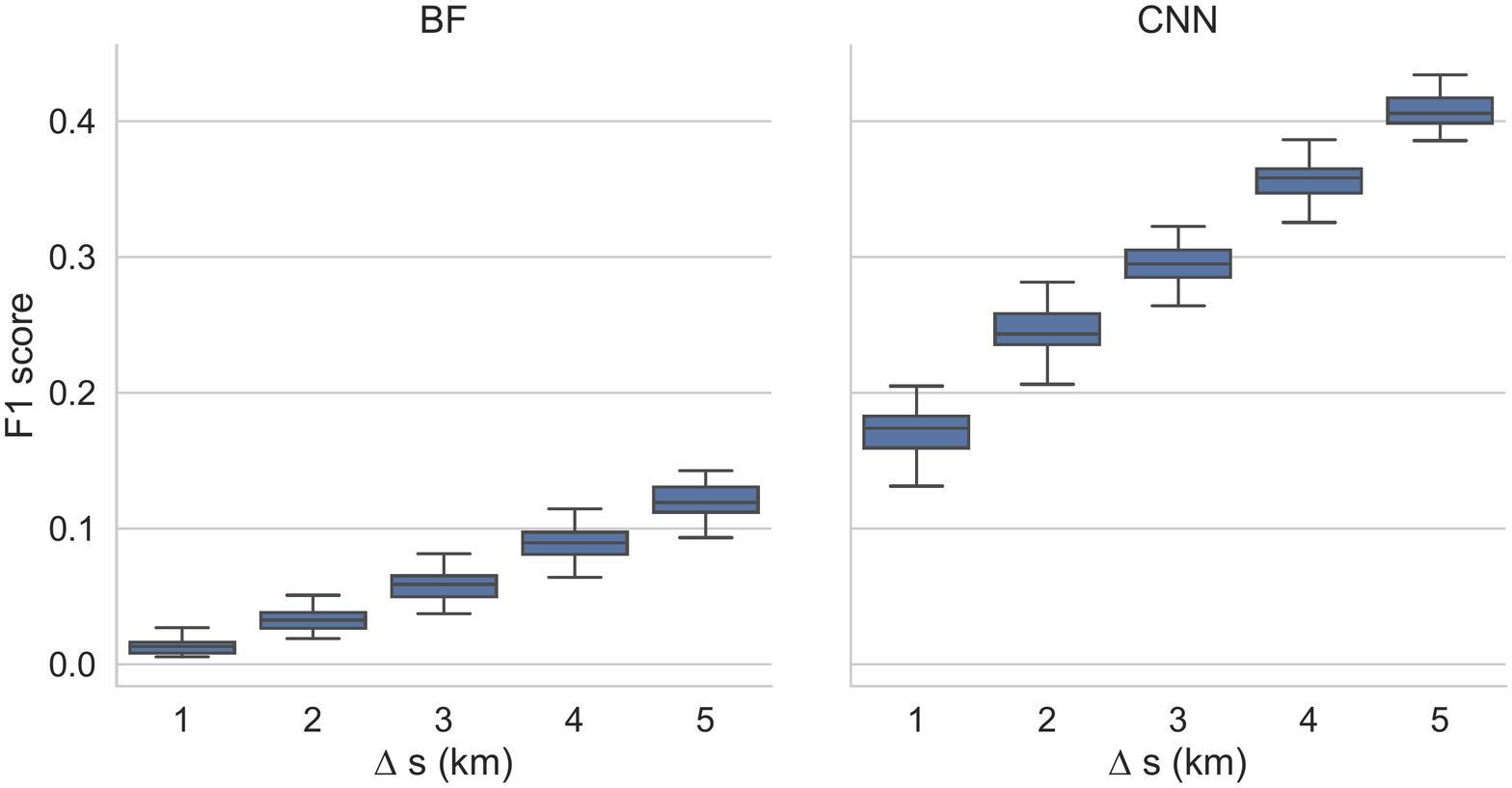}
    \caption{Performance of models for different $\Delta s$.}
    \label{fig:performance-by-delta-s}
\end{subfigure}
\begin{subfigure}[b]{0.49\textwidth}
    \centering
    \includegraphics[trim={2.5cm 4cm 0 2cm},clip,width=\textwidth]{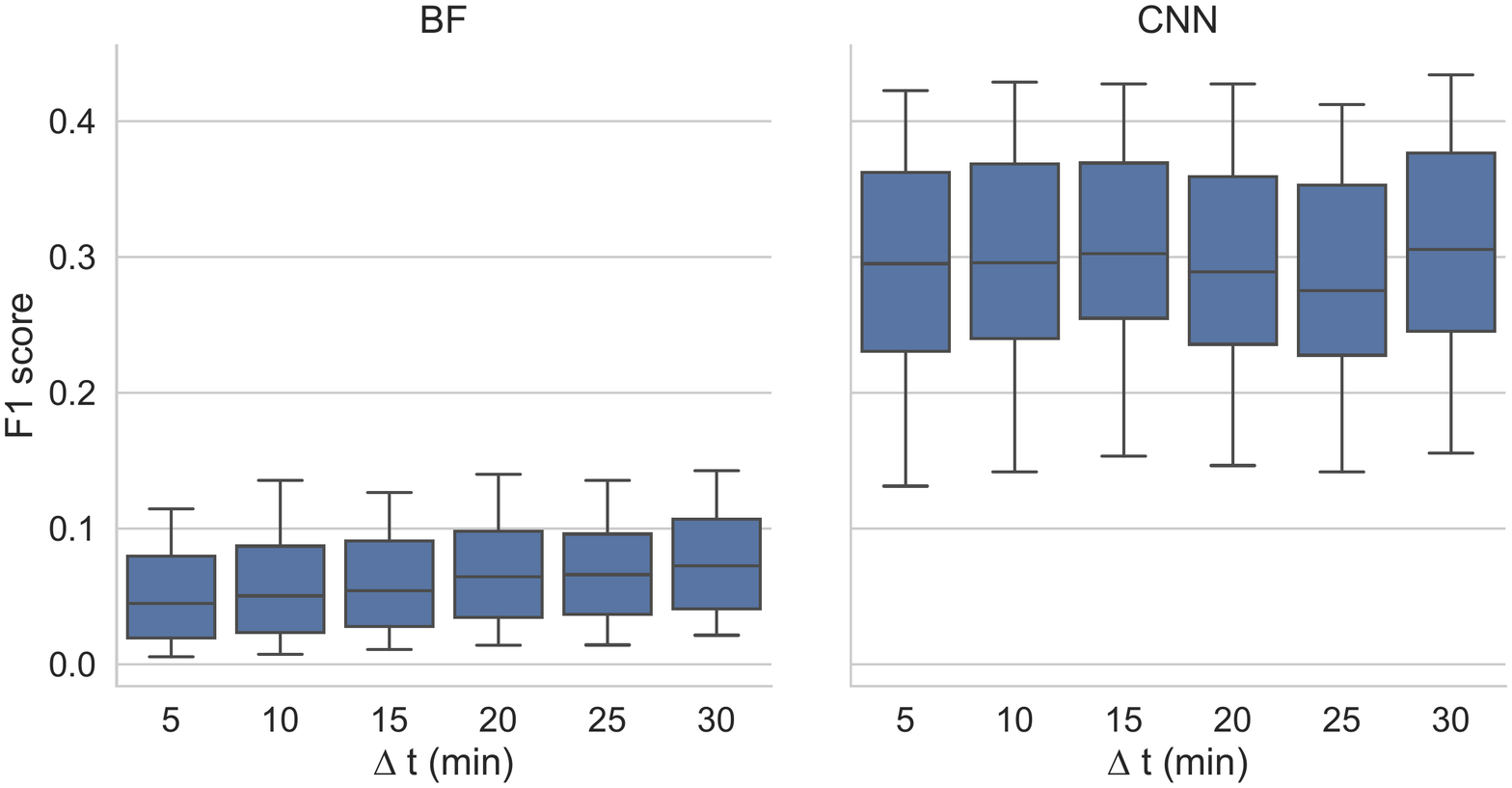}
    \caption{Performance of models for different $\Delta t$.}
    \label{fig:model-performance-by-delta-t}
\end{subfigure}
\caption{Performance of models against different spatial and temporal resolutions using the test data sets. We observe that the CNN model (part of CROME) significantly outperforms the baseline approaches.}
\label{fig:model-performance-by-delta-x}
\end{figure*}

\subsection{Solution Approach}
\label{sec:proposed_approach}

\subsubsection{Optimizing Detection Capability}
Our goal is to solve optimization problem~\ref{eqn:optimization}. We begin by describing the function $f$. An approach to detecting incidents must be able to balance between the precision and the recall of detection. First responders typically work under resource constraints; while early detection is imperative to minimizing response times, responders cannot be dispatched to attend to falsely generated alerts. As a result, we choose to use F-1 score as our metric of interest, which is the harmonic mean of precision and recall. Given our choice of $f$, we now focus on the first term in Formula ~\ref{eqn:optimization}, which seeks to find the parameters $\theta^{*}$ such that $\theta^{*} = \text{argmax}_{\theta} f(I \mid \mathbb{X}, \theta)$. 

We leverage the structure of the incident detection problem while choosing a model that can draw inferences from spatial-temporal crowdsourced data. Intuitively, we try to capture the proximity of observation and reporting in this domain. After users observe an incident, they move in space and a certain amount of time elapses before they report it. However, it is natural to assume that the displacement (both spatial and temporal) is not very high; it is unlikely that users move many miles and report an incident hours after they observe it. Therefore, reports generated in a cell (say $g_i \in G$) are most likely to have been observed in cells that lie in close proximity to $g_i$. In principle, it is possible to flatten $\mathbb{X}$ to a vector and minimize some loss function (e.g., least square loss) that measures prediction accuracy. However, to leverage the spatial structure of the problem, we use a convolutional neural network (CNN)~\cite{goodfellow2016deep} to maximize $f$. We implement two convolutional layer neural network with a max-pooling layer in between. The last convolutional layer is followed by a ReLU activated layer and a sigmoid layer. The CNN architecture enables us to consider how alerts generated in a cell are correlated with incidents that occur in its spatial proximity. 

\subsubsection{Optimizing Practitioner Parameters}
Our goal is to maximize the accuracy of detection while also improving spatial and temporal localization. The very structure of our formulation dictates that no single solution can optimize all the objectives simultaneously. Indeed, we show through experimental evaluation that as the spatial resolution increases, the accuracy of the model degrades. As a result, we seek to find the set of Pareto-optimal solutions. A solution is called Pareto-optimal or non-dominated if none of the objective functions can be improved without sacrificing the value of at least one of the other objective  functions~\cite{deb2005evaluating}. 
To find the Pareto optimal solution, we use a multi-objective evolutionary algorithm~\cite{deb2005evaluating} 
based on the concept of $\epsilon$-dominance~\cite{laumanns2002combining}. The idea of $\epsilon$-dominance maintains a well distributed set of non-dominated solutions by not allowing two solutions with a difference less than an exogenously specified threshold ($\epsilon_i$ in the $i$th objective) to be non-dominated to each other. We refer readers to work by Deb et al.~\cite{deb2005evaluating} for a detailed description of 
multi-objective optimization. 

\subsubsection{Feature Generation} 
We collect a set of crowdsourced reports about roadway accidents through the Waze application. We remove duplicate reports that had the same identifier (ID). 
Then, we generate input-output pairs $\{\mathbb{X}, I\}$ based on the spatial discretization, temporal discretization, report aggregation, and label assignment as described in section~\ref{sec: Problem_Statement}.
We use three major categories of features for incident detection -- crowdsourced reports, traffic information, and weather. Our choice is guided by prior work in the domain of accident prevention and analysis~\cite{mukhopadhyay2020review}. To capture the volume as well as the reliability of the crowdsourced reports, we generate the following three features: 
1) Volume: the total number of crowdsourced reports, 2) Sum of Reliability:  the sum of the reliability scores of the crowdsourced reports, and 3) Mean of Reliability: the average reliability score of the crowdsourced reports. We gather reliability scores for the reports from the crowdsourcing platform (Waze) itself. 

We generate two additional features from traffic and weather information. For a point in space and time (based on the spatial-temporal discretization), we gather the amount of precipitation from the nearest weather station. We also calculate the mean traffic congestion in a cell through a two-step process. First, for each roadway segment in a cell, we calculate congestion by computing the ratio of the difference between free flow speed and the current speed to free flow speed. Then, we average the value of congestion across all the roadway segments within the cell. 

\section{Data}
\label{sec:data_platforms}

We use the following data sources for learning the incident detection model:
\begin{enumerate*}
\item \textbf{Crowdsourced data}: Waze is a GPS navigation application and crowdsourcing platform~\cite{waze_driving_2020}. We look at user reports concerning roadway accidents from Sep/01/2019 to Dec/31/2009;
\item \textbf{\textit{Ground-Truth} Incident Data}: we collect accident data from the public safety office of Nashville, USA, with a size of about 500 sq. miles  We consider such incidents as the ground truth. To remove noise, we map each incident to its closest roadway segment (typically at a distance of less than 25m); 
\item \textbf{Traffic Data}: we collect roadway traffic data in a time resolution of 5-minute intervals for the area under consideration, resulting in approximately 270 million measurements; and 
\item \textbf{Weather Data}: we collect weather information from Weatherbit~\cite{weatherbit}.
\end{enumerate*}

\section{Experimental Evaluation}
\label{sec:Experimental_Evaluation}

\subsection{Setup, Baseline, and Implementation}
As indicated in Section~\ref{sec:data_platforms}, we collect crowdsourced data from the Waze platform for the period between Sep/01/2019 to Dec/31/2019. We divide the data between training sets of three months and a test set of one month in a manner that each month is used as the test set in an independent evaluation. Hyper-parameters of 
the CNN (number of epochs and threshold for classification) are tuned through $k$-fold cross-validation. 
We use 256 filters with filter size of $2 \times 2$ for both convolutional layers. Moreover, max~pooling size is set to $2 \times 2$. Last two dense layers contain $2 \times N_o$ and $N_o$ units accordingly.
Moreover, we set $\alpha$ and $\beta$ equal to 1 hour and $T^\prime$ equal to 30 minutes based on empirical analysis. 
To compare the performance of CROME, we use our earlier implementation using Bayesian Information Fusion (BF) shown in \cite{senarath2020emergency}.
Our implementation for CROME as well as the baseline approach is available at \url{https://github.com/ysenarath/CROME}.

\subsection{Evaluation Measures}
\label{subsection:Evaluation Measures}

In order to evaluate the performance of the proposed approach, we use four metrics chosen based on the priorities of first responders and practitioners. Specifically, we evaluate using \textbf{F-1 score} (harmonic mean of precision and recall), \textbf{average early prediction ratio} (the ratio of correctly early-predicted incidents to total incidents reported), \textbf{average early prediction distance} (the geodesic distance between the detected location of the incident and the actual location), and the \textbf{average early prediction time} (the average time difference between the actual reported time of the incident obtained through ground-truth report and the time of detection).

\begin{table*}[ht]
\small
\centering
\caption{Early detection performance of models. The F1 score is evaluated on the test set.}
\label{tab:early-detection-performance-2}
\resizebox{0.9\textwidth}{!}{%
\begin{tabular}{@{}lrrrrrrrr@{}}
\toprule
\textbf{Model} & \multicolumn{1}{l}{\textbf{$\Delta$s (km)}} & \multicolumn{1}{l}{\textbf{$\Delta$t (min)}} & \multicolumn{1}{l}{\textbf{F1 Score}} & \multicolumn{1}{l}{\textbf{Early Pred \%}} & \multicolumn{1}{l}{\textbf{Avg Distance (km)}} & \multicolumn{1}{l}{\textbf{\begin{tabular}[c]{@{}l@{}}Avg. Early\\ Time (min)\end{tabular}}} & \multicolumn{1}{l}{\textbf{Precision}} & \multicolumn{1}{l}{\textbf{Recall}} \\ \midrule
\multicolumn{9}{c}{\textbf{Best Early Pred \%}} \\
BF & 1 & 5 & 0.60 & \textbf{77.56} & 3.27 & 15.02 & 0.00 & 0.14 \\
CROME & 5 & 5 & \textbf{41.00} & 40.28 & 2.96 & 13.94 & 0.32 & 0.56 \\ \midrule
\multicolumn{9}{c}{\textbf{Best Avg. Distance}} \\
BF & 5 & 5 & 10.56 & 35.00 & 3.16 & 15.02 & 0.06 & 0.33 \\
CROME & 1 & 5 & \textbf{16.41} & 18.35 & \textbf{2.67} & 14.32 & 0.16 & 0.18 \\ \midrule
\multicolumn{9}{c}{\textbf{Best Avg. Early Time}} \\
BF & 3 & 20 & 6.49 & 47.69 & 3.25 & \textbf{15.45} & 0.04 & 0.40 \\
CROME & 3 & 30 & \textbf{30.40} & 24.67 & 3.03 & 14.92 & 0.23 & 0.45 \\ \bottomrule
\end{tabular}%
}
\vspace{-0.05in}
\end{table*}


\subsection{Results}

\subsubsection{Detection Accuracy}
We begin by comparing the detection accuracy of the CNN model (the first component of CROME) with respect to the baseline approaches. We vary the spatial and temporal resolutions ($\Delta s$ and $\Delta t$) for this comparison and present the results without using the Pareto optimization framework (we present complete results later). Our purpose of doing so is two-fold. We seek to examine the robustness of the models with respect to varying discretization parameters and also validate our hypothesis that detection accuracy can suffer as spatial and temporal resolutions increase. We present the results in Figure~\ref{fig:performance-by-delta-s} and Figure~\ref{fig:model-performance-by-delta-t} that show the influence of time and space resolution on F1 score. We have the following major findings: \textbf{1)} CROME significantly outperforms the baseline models in terms of F-1 score in all cases. \textbf{2)} As the spatial resolution increases and localization becomes more challenging, the F-1 score of all the models decrease. \textbf{3)} Surprisingly, temporal discretization does not affect the performance of any of the models (barring minor variations). We hypothesize that this is due to the effects of aggregating features over multiple time steps ($T^\prime$).

\begin{figure}[t]
\vspace{-0.1in}
\centering
\includegraphics[trim={5cm 20cm 2cm 20cm},clip,width=\columnwidth]{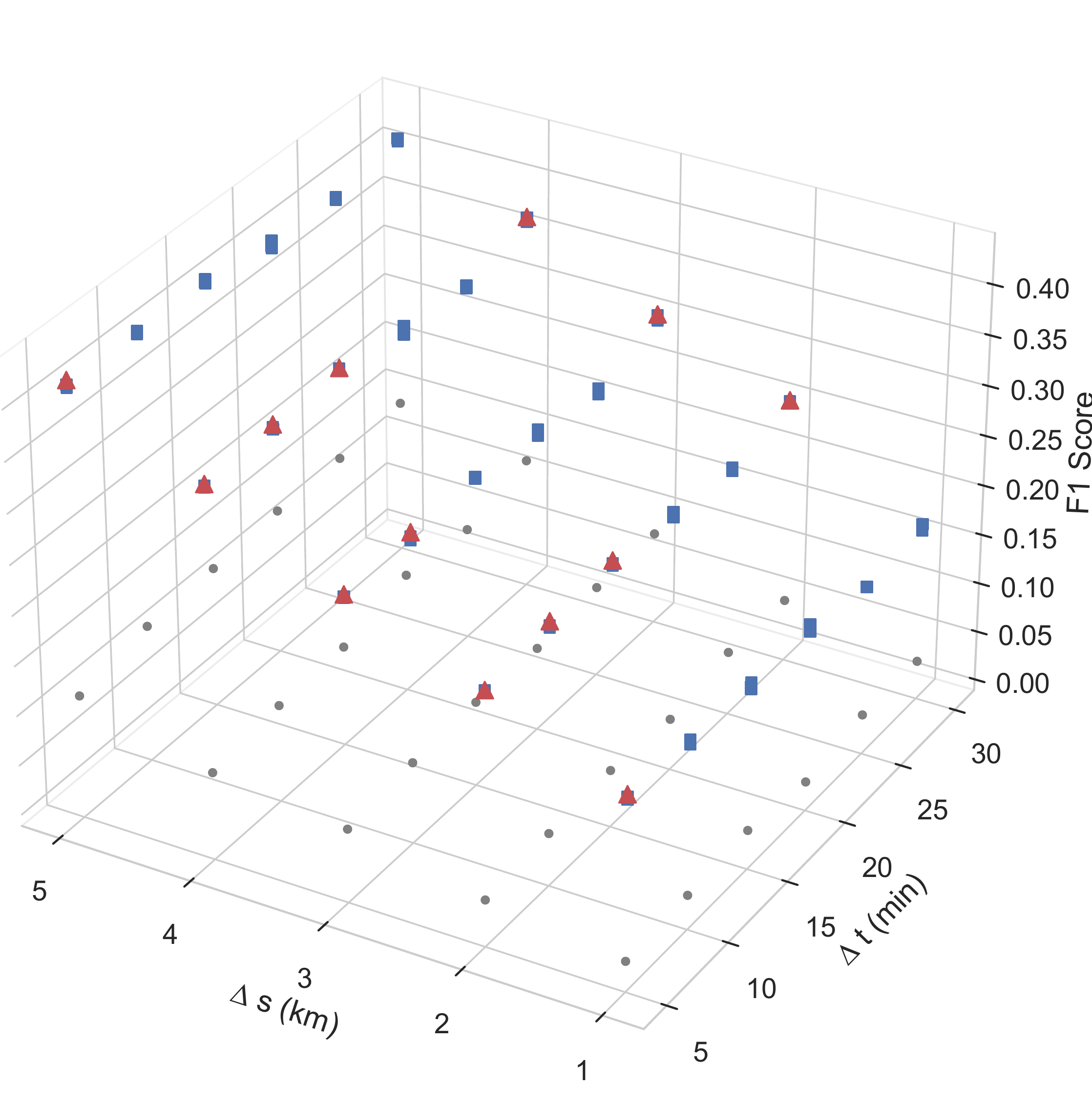}
\caption{We show the non-dominated set of solutions obtained through CROME (in red triangles). The learned CNN models are shown in blue squares. We also show the baseline models (in gray circles).} 
\label{fig:pareto-optimal-solutions}
\end{figure}

\subsubsection{Choosing the Pareto Frontier}
Having shown that the CNN model we propose as part of CROME outperforms the baseline approaches (in terms of F-1 score), we now focus on evaluating CROME in its entirety. Recall that our goal is to simultaneously aid detection and localization by solving optimization problem~\ref{eqn:optimization}. We first show how calculating the Pareto frontier helps us solve optimization problem~\ref{eqn:optimization} (see Figure~\ref{fig:pareto-optimal-solutions}). The plot shows the three dimensions over which CROME optimizes, namely the spatial resolution ($\Delta s$), the temporal resolution ($\Delta t$), and the F-1 score. Each point in the three-dimensional plot represents a specific \textit{learned} model. We show the non-dominated set in red triangles. Notice that the non-dominated set consists of several learned models. We leave the final choice of selecting one (or more) models for deployment from the non-dominated set to the practitioners based on the relative importance of the specific objective functions ($f, \Delta s,$ and $\Delta t$). 



Our key findings are as follows: \textbf{1)} While BF outperforms CROME in terms of early detection, $85\%$ of the alerts that it generates are incorrect (false positives). Using such an approach is infeasible in practice as responders cannot be dispatched based on incorrect alerts. CROME on the other hand, balances precision and recall to detect more than $40\%$ of the incidents early. \textbf{2)} CROME performs nearly on-par with the baseline approach in terms of localization. 
\textbf{3)} We find that while tailoring models to specifically maximize precision or recall can maximize certain metrics associated with incident detection, the lack of consideration of practitioner-specific parameters can lead to detrimental consequences. 
We note 
the need to consider a combination of F-1 score and the metrics pertaining to detection and localization. Consider a model that generates alerts at all time-steps on all cells. Such a model would detect all possible incidents early with remarkable localization (since localization is measured with respect to the ground-truth incidents). As a result, the balance of precision and recall is crucial in this setting. Table~\ref{tab:early-detection-performance-2} shows the results of our experiments. We have selected the best model of each approach (BF, CROME) with respect to three practitioner-centric metrics. The model is considered better if there is a higher early prediction rate, higher avg. early prediction time, or lower avg. distance.
Based on Table~\ref{tab:early-detection-performance-2}, we conclude that the proposed approach, which seeks to include practitioner-centric parameters in incident detection, results in a significantly higher F-1 score, significantly lower number of false alerts, and competitive spatio-temporal localization. 
\section{Conclusion}

This paper proposes a novel multi-objective optimization problem for early detection of spatial-temporal incidents using crowdsourced data. We show how crowdsourced data, historical ground-truth incident data, and relevant determinants can be combined to detect potential incidents before they are reported. We also show how practitioner-centric parameters can be incorporated into our approach. The proposed approach, CROME (Crowdsourced Multi-objective event detection), uses a combination of convolutional neural networks and Pareto optimization to solve the optimization problem. 
While crowdsourced data is often noisy and uncertain, 
we show that mining such non-traditional data can benefit emergency incident response through a principled modeling approach like CROME. 

\spara{Acknowledgement.} This project was supported by resources provided by the Office of Research Computing at George Mason University and funded in part by the National Science Foundation grants (1814958, 1815459, 1625039, 2018631) and a grant from Tennessee Department of Transportation. 


\bibliographystyle{IEEEtran}
\bibliography{bibliography}
\end{document}